# Spatio-Temporal Graph Neural Networks: A Survey

Zahraa Al Sahili, *

Queen Mary University of London, z.alsahili@qmul.ac.uk

Mariette Awad

American University of Beirut, ma163@aub.edu.lb

Graph Neural Networks have gained huge interest in the past few years. These powerful algorithms expanded deep learning models to non-Euclidean space and were able to achieve state of art performance in various applications including recommender systems and social networks. However, this performance is based on static graph structures assumption which limits the Graph Neural Networks' performance when the data varies with time. Spatio-temporal Graph Neural Networks are extension of Graph Neural Networks that takes the time factor into account. Recently, various Spatio-temporal Graph Neural Network algorithms were proposed and achieved superior performance compared to other deep learning algorithms in several time dependent applications. This survey discusses interesting topics related to Spatio-temporal Graph Neural Networks, including algorithms, applications, and open challenges.

## 1 INTRODUCTION

Graph Neural Networks (GNNs) are a class of deep learning models that are specifically designed to operate on graph-structured data. These models leverage the graph topology to learn meaningful representations of the nodes and edges of the graph. GNNs are an extension of traditional convolutional neural networks, and they have been shown to be effective in tasks such as graph classification, node classification, and link prediction. One of the key advantages of GNNs is their ability to maintain good performance even as the size of the underlying graph grows, thanks to the independence of the number of learnable parameters from the number of nodes in the graph.

GNNs have been extensively used in various domains such as including recommendation systems, drug discovery and biology, and resource allocation in autonomous systems. However, these models are limited to static graph data, where the graph structure is fixed. In recent years, there has been an increased interest in time-varying graph data, which appears in various systems and carries valuable temporal information. Applications of time-varying graph data include include multivariate time series data, social networks, and audio-visual systems.

To address this need, a new family of GNNs has emerged: spatio-temporal GNNs, which takes into account both the spatial and temporal dimensions of the data by learning temporal representations of the graph structure. In this survey, we provide a comprehensive review on the state-of-the-art spatio-temporal GNNs. We start by giving a brief overview of the different types of spatio-temporal GNNs, and their underlying assumptions. We then delve into the specific algorithms used in spatio-temporal GNNs in greater detail, while also providing a useful taxonomy for grouping these models. We also provide an overview of the various applications of spatio-temporal GNNs, highlighting the key areas where these models have been used to achieve state-of-the-art results. Finally, we discuss open challenges and future research directions in the field.

---

* corresponding author

In summary, this survey aims to provide a comprehensive and in-depth look at spatio-temporal GNNs, highlighting the current state of the field, the key challenges that still need to be addressed, and the exciting future possibilities for these models.

## 2 BACKGROUND

Graph Neural Networks (GNNs) are a class of neural networks that operate on graph-structured data. They are designed to learn from and make predictions on graph-structured data, such as social networks, molecular structures, and transportation networks. GNNs can be broadly classified into two categories: spectral and spatial GNNs.

Spectral GNNs make use of the eigenvectors and eigenvalues of the graph Laplacian matrix to define the graph convolutional operation. A popular example of a spectral GNN was presented in [2], which utilizes the graph Laplacian to define a convolutional operation applied to the node features. On the other hand, Spatial GNNs use the graph structure to define the convolutional operation and typically involve updating the node representations based on the representations of its neighboring nodes. Examples of Spatial GNNs include Graph Attention Network (GAT)[3] and Graph Isomorphism Network (GIN)[4]. GAT uses attention mechanisms to weigh the importance of different neighbors when updating node representations, while GIN is an extension of GCN that uses a multi-layer perceptron to update node representations, rather than using the graph Laplacian.

Recently, there has been a significant advancement in GNNs with the introduction of Graph Transformer [5], which is an extension of the transformer architecture to operate on graph-structured data. The Graph Transformer uses self-attention mechanisms to update node representations and has been shown to achieve state-of-the-art results on several graph-based tasks.

In conclusion, GNNs are a powerful tool for working with graph-structured data and have demonstrated great potential in a wide range of tasks. The choice of GNN architecture depends on the specific task and the properties of the graph data. Spectral GNNs like GCN are better suited for tasks that involve node classification and link prediction, while spatial GNNs like GAT and GIN are more suitable for tasks that involve graph classification and node clustering. With the recent development of Graph Transformer, it has also shown great potential in graph-based tasks.

## 3 ALGORITHMS

Spatio-temporal graph neural networks can be classified from algorithmic perspective as spectral based and spatial based. Another classification category is the method time variant was introduced: wether using another machine learning algorithm or defining time within the graph structure.

**3.1 Hybrid Spatio-Temporal Graph Neural Networks**

Hybrid Spatio-temporal graph neural networks constitute of two main components: a spatial component and a time component (Figure1) .

*3.1.1Spatial Module*

In hybrid Spatio-temporal graph neural networks, graph neural network algorithms are used to model the spatial dependencies in the data.

### 3.1.1.1 Spectral Graph Neural Networks

Spectral GNNs are based on the spectral definition of convolution operation. Early Spatio-temporal GNNs heavily relied on this spectral definition. For example, Yu et al. , used Chebyshev GNN in the STGCN algorithm [6]. In addition, Cao et al. used Spectral graph convultion to model the space domain in his StemGNN [7]. Recently, Simeunivic et al. used spectral GCNs in his both algorithms: GCLSTM and CGTransfo [8].

### 3.1.1.2 Spatial Graph Neural Networks

With the advances in spatial graph neural networks research, various researchers used spatial GNNs to model the spatial domain in spatio-temporal GNNs.Chen et al[9], used recurrent graph neural network(RGNN) with skip connection to model spatial dependencies in traffic forecasting. However, Wu et al. used Graph Convolution neural networks GCNs with skip connection in his MTGNN algorithm [10]. Additionally, GCN was used in the Structural RNN algorithm [11].

Graph neural networks with attention mechanism (GAT) was used in [12,13]. In A2GNN, Huang et al. used the GAT with auto graph learner to improve the forecasting performance [2]. Moreover, Kan et al. used GAT cascaded with a graph transformer and a heriraical pooling mechanism in the HST-GNN implementation [13].

More advanced spatiotemporal GNN algorithms were used for space modelling in [14,15]. Oreshkin et al. used Gated graph neural networks in the FG-GAGA algorithm. In contrast, Graph Isomorphism network was used by Kim et al. to model brain connectivity in brain graph representation.

### 3.1.1.3 Graph Transformers

Two algorithms relied on graph transforms to model spatial dependencies: TransMOT and Forecaster [16,17].In addition, Kan et al. accompanied his GAT with a Graph transformer in his HST-GNN architecture [13].

*3.1.2 Temporal Module*

To model the time domain, various machine learning algorithms can be involved.

### *3.1.2.1 1D-CNN*

Yu et al. used a 1D-CNN to account for the time domain in his STGCN algorithm. Moreover, Wu et al [10], used an inception layer in the MTGNN implementation[6]. Also, Cao et al. used 1D CNNs with GLU units for the temporal module [7].

### *3.1.2.2 Recurrent Neural Networks*

Recurrent neural networks and its variants as Gated Recurrent units (GRUs) and Long Short Terms Memory units (LSTMs) were widely adopted in hybrid spatio-temporal GNNs to model the time domain. Jain et al used RNNs in the structural RNN algorithm [11]. On the other side, Oreshkin et al. used GRUs in FG-GAGA GNN while Chen et al. [9] used both GRUs and LSTMs in the MResGNN algorithm. In addition, [8,12,13,18] all used LSTMs as time modules. In the HST-GNN algorithm, Kan et al. used 2 LSTMs with an attention mechanism within a wider encoder decoder architecture [13].

### *3.1.2.3 Transformers*

Recently, a huge focus was imposed on transformer architectures that were used to accompany for the time domain. Transformer was used by [8,15,16,17] in the TransMOT, Forecaster, STAGIN, and GCTransfo respectively.

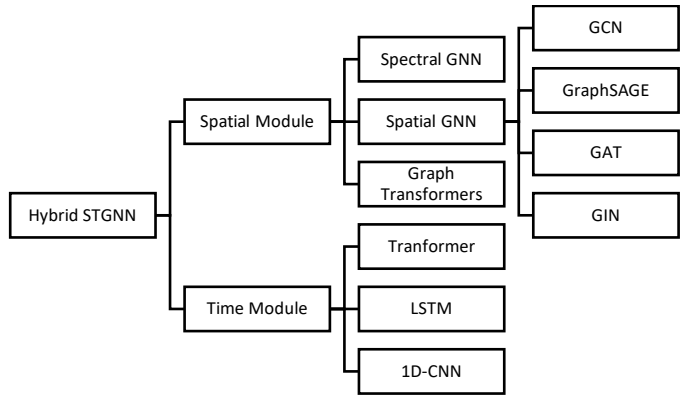

Fig. 1. Hybrid based Spatio-Temporal Graph Neural Networks

**3.2 Solo-Graph Neural Networks**

Another method to model time in spatio-temporal graph neural networks is to define the temporal frame within the GNN itself. Multiple approaches were proposed including: defining time as an edge, inputting time as a signal to the GNN, time modeled as a subgraph, and sandwiching other machine learning architectures inside the GNN (Figure 2).

*3.2.1 Time as Edge*

Kapoor et al. [19] used spatial GCN with skip connections to forecast covid. In this algorithm, time was defined as an edge and locations as graph nodes. Additionally, time was defined as an edge in USTGCN algorithms which modified the space adjacency matrix to a space-time adjacency matrix [20].

*3.2.2 Time as Signal*

Time as an input signal was used in GNN pure based spatio-temporal GNNs. Zhang et al. used temporal hierarchy modelling to input time to the GAT [21]. The algorithm also included graph trimming and convolution diffusion to improve the performance. Moreover, Shen et al. used a gated dilated casual block for the temporal input [22]. The output of this block was inputted to a dynamic GCN in parallel with the output of a similar double block for the spatial domain. Time was also inputted as a signal in the CasualGNN algorithm [23]. The algorithm is based on dynamic graph neural network with attention mechanism and a casual module.

*3.2.3 Time as Subgraph*

Li et al modelled time as a subgraph within a graph isomorphism network (GIN) [24]. Moreover, Shao et al. used temporal similarity graph to account for the temporal domain, which was added to other spatial graphs to form a multigraph set that constructed the ASTGCN framework [25].

*3.2.4 Time using Sandwiching*

Karimi et al. used two 1D-CNNs to model time. In this architecture, the 1D-CNNs were sandwiched inside the GCN architecture as sub modules [11].

*Time as Filter*

In Space time Graph Neural Network, both time and space were introduced as multivariate integral Lipschitz filters inside the GCN [26].

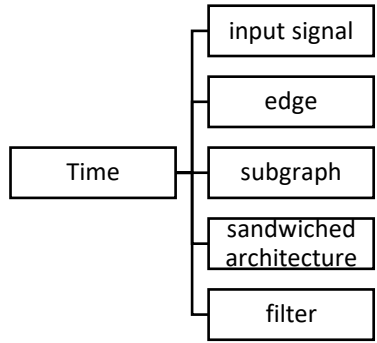

Fig. 2. Time Modelling in GNN only algorithm

| Author | Name | Hybrid Algorithm | GNN only | Spectral Based | Spatial Based | Spatial Module | Time Module |
|---|---|---|---|---|---|---|---|
| Yu et al. | STGCNGCN | √ | | √ | | Chebychev GNN | 1D-CNN |
| Nicolivioiu et al. | RSTGCN | √ | | | √ | Custom(3D CNN) | LSTM |
| Chen et al. | MResGNN | √ | | | √ | RGNN | GRU & LSTM |
| Kapoor et al | - | | √ | | √ | GCN; time added as an edge | |
| Wu et al. | MTGNN | √ | | | √ | GCN | Inception layer |
| Li et al. | Unified GNN | | √ | | √ | GIN; time added as a subgraph module | |
| Cao et al. | StemGNN | √ | | √ | | Spectral GCN | 1D-CNN |
| Oreshkin et al. | FG-GAGACN | √ | | | √ | GGCN | GRU |
| Jain et al. | Structural RNN | √ | | | √ | | RNN |
| Karimi et al. | St-GNN | | √ | | √ | GCN with 1DCNN sandwiched in the GNN | |
| Chu et al. | TransMOT | √ | | | √ | Graph transformer | Transformer |
| Kim et al. | Forecaster | √ | | | √ | Graph Transformer | Transformer |
| Kim et al. | STAGIN | √ | | | √ | GIN | Transformer |
| Zhang et al. | ST-GDN | | √ | | √ | GAT;time as input using temporal heriracly modeling | |
| Shao et al. | ASTGCN | | √ | | √ | Multi graph set;Time as Temporal similarity graph | |
| Huang et al. | A2GNN | √ | | | √ | GAT | LSTM |
| Simeunovic et al. | GCLSTM | √ | | √ | | Spectral GCN | LSTM |
| Simeunovic et al. | GCTransfo | √ | | √ | | Spectral GCN | Transformer |
| Kan et al. | HST-GNN | √ | | | √ | Graph Transformer &GAT | LSTM |
| Hadous et al. | Space Time GNN | | √ | | √ | GCN; time as a filter | |
| Shen et al. | T2GNN | | √ | | √ | Dynamic GCN: time as input signal | |
| Wang et al. | CasualGNN | | √ | | √ | GAT;time as a signal | |
| Roy et al. | USTGCN | | √ | | √ | GCN;time as an edge | |

*Table I.STGNN Algorithms*

# 4 APPLICATIONS

## 4.1 Multivariette Time series Forecasting

Motivated by the power of GNNs in handling relational dependencies [10], spatio-temporal GNNs were widely applied in multivariate time series forecasting. Applications include traffic forecasting, Covid forecasting, PV power consumption, RSU communication, and seismic applications.

*4.1.1Traffic*

Transportation is considered a very important factor in every person's life [6]. Based on a study conducted in 2015, U.S . drivers spend a daily average of 48 minutes behind the wheel [27]. Thus, accurate real-time forecast of traffic conditions is of dominant importance for road users, private sectors and governments. However, traditional machine learning forecast systems fails to satisfy accuracy conditions due to the high nonlinearity and complexity of traffic flow [6]. In contrast and based on the power of GNNs in handling non linearities, spatiotemporal graph neural networks were widely applied in traffic forecasting in both aspects: long term and short-term predictions [6,7,9,12,14,17,20,21,22,24].

*4.1.2Pandemic Forecasting*

In a state of pandemic, the ability to accurately forecast caseload is extremely important to country level or the individual level [19]. With conventional algorithms considering forecasting pandemic cases as a closed loop based on previous cases and considering the spatial dependencies between neighborhoods in effecting pandemics, spatio-temporal graph neural networks were used to accompany for both space and time in pandemics. Several spatio-temporal graph neural network algorithms were proposed and found to achieve state of art COVID forecasting in the United States, United Kingdom, Germany, and worldwide [7,10,19,23]

*4.1.3PV*

Due to the rapid increase in installation of commercial PV power plants, the operation and planning for reliable performance of PV systems is a crucial challenge [28].Ensuring reliable performance includes monitoring the slow loss of electricity output and effective planning based on the PV power output. This reliability can be achieved by accurate power forecasting. Based on the ability of GNNs in capturing spatial and temporal dependencies, spatio-temporal graph neural networks were widely adopted to forecast PV power [7.8,11] and were able to achieve superior performance over other forecasting algorithms.

*4.1.4RSU communication*

As a special type of base station, Road Side Units (RSU) can be deployed at low cost and effectively alleviate the communication burden of regional Vehicular Ad-hoc Networks (VANETs) [29]. Unfortunately, due to the limited energy storage and peak hour communication demands in VANETs, it is crucial for RSUs to adjust their participation in communication according to the requirements and allocate energy reasonably to balance the workload. Zheng et al. proposed a spatio-temporal graph neural network algorithm the forecast RSU network load through inputting the historical information around RSU and the topological relationship between RSU [29].

*4.1.5 Seismic*

Bloemheuvel et al. used spatio-temporal graph neural networks to predict earthquakes. The proposed approach achieved superior performance over other traditional machine learning algorithms.

**4.2 Human Object Interaction**

Learning in the space-time domain remains a very challenging problem in machine learning and computer vision [18]. The main challenge is how to model interactions between objects and higher level concepts, within the large spatio-temporal context[18]. In such a difficult learning task it is critical to efficiently model the spatial relationships, the local appearance, and the complex interactions and changes that take place over time. Nicolicioiu et al. introduced a spatio-temporal graph neural network model, recurrent in space and time, suitable for capturing both the local appearance and the complex higher-level interactions of different entities and objects within the changing world scene [18].

**4.3 Dynamic graph representation**

Temporal graph representation learning has been considered a very important aspect in graph machine learning [15,31]. With limitations of existing methods in capturing powerful representations due to reliance on discrete snapshots of the temporal graph, Wen et al. proposed a dynamic graph representation learning method using spatio-temporal graph neural networks [31]. Moreover, Kim et al. used spatio-temporal GNNs to dynamically represent brain graphs [15].

*Multiple object tracking*

In addition, tracking multiple objects in videos heavily depend on on modeling the spatial-temporal interactions between objects [16]. Chu et al. proposed a spatio-temporal graph neural network algorithm that models spatial and temporal interactions among the objects [16].

**4.4 Sign Language Translation**

Sign languages, which engage visual-manual modalities to convey meanings, are the primary communication tools for the deaf and hard-of-hearing community [13]. To reduce the communication gap between spoken language and sign language users, machine learning is involved. Traditionally, neural machine translation has been heavily adopted while more advanced methods are needed to capture the spatial properties in sign languages. Kan et al. presented a spatio-temporal graph neural network-based translation system, that is powerful in capturing spatial and temporal structures of sign language which led to state of art performance compared to traditional neural machine translation methods [13].

**4.5 Techology growth ranking**

Understanding the growth rate of technologies is a core key in technology sector business strategy [32]. In addition, predicting the growth rate of technologies and the relations to each other informs business decision-making in terms of product definition, marketing strategies, and research and development [32]. Cummings et al. proposed a methodology to predict technology growth ranking from social networks using spatio-temporal graph neural networks [32].

**4.6 Knowledge Graphs and Social Networks**

Real world graphs like social networks and knowledge graphs are dynamic. For example, in a social network, new users join over time and users interact with each other through messages and posts reactions[33]. In addition, new events appear with time in knowledge graphs. To account for the evolving dynamic properties in graphs, [33] introduced a temporal graph neural networks that can handle billions of nodes and edges and can jointly learn the temporal, structural, and contextual relationships on dynamic graphs.

### 4.7 Audio Visuals and Emotion Perception

Effective dimension prediction from multi-modal data is becoming an increasingly challenging and important research area [33]. For example, discriminative features from multiple modalities are critical to accurately recognize emotional states. Motivated by their spatial and temporal power, Chen et al. investigated spatio-temporal graph neural networks in audio visuals [33]. The framework achieved superior performance compared to traditional deep learning frameworks when experimented on emotional recognition applications. In addition, Bhattacharya et al. proposed a spatio-temporal graph neural network that leverages emotion perception [34].

## 5 OPEN CHALLENGES

### 5.1 Automation

Automation in machine learning is an important research topic. AutoML, is a promising direction to improve the algorithms performance and scale up the models training. With available techniques that are explored in neural networks, expansion of this research area to spatio- temporal graph neural networks is a promising direction.

### 5.2 Benchmarking

Spaio-temporal algorithms lack benchmark datasets. Every proposed spatio-temporal graph neural network algorithm is trained on a custom dataset and not compared to compartment models. This might be caused by the lack of benchmarks which make it time consuming to retrain all compartment models. Presenting a benchmark for spatio-temporal data is very important in understanding the algorithms and comparing their performance.

### 5.3 Augmentation

Neural Networks heavily rely on large data. As neural network extension, spatio-temporal graph neural networks need abundant data for superior performance. Wang et al. attempted to augment temporal graphs using MeTA (Memory Tower Augmentation) module that provide adaptively augmented inputs for every prediction. Nevertheless, augmenting spatio-temporal data is still considered a very challenging task which advances in it can lead to huge improvement in this research era [35].

### 5.4 Privacy/Federated Learning

Privacy of users is an ethical concern. Federated learning methods are the leading approach to privacy preserving algorithms. With a single federated learning research attempt, more approaches are critically needed in the spatio-temporal graph neural networks era.

### 5.5 Pretraining and Transfer Learning

Neural networks rely of abundant data to perform efficiently. In scarce data setting, transfer learning is considered the most affordable approach to achieve high performance. In spatio-temporal graph neural networks, transfer learning is still a challenge. Lou and Panagopoulos et al. proposed transfer learning frameworks for Covid forecasting and Highway traffic modeling and forecasting approaches are critical for intelligent transportation systems [36,37]. Advances in this field include investigating transferability of spatio-temporal graph neural networks added to proposing similarity metrics and methods that mitigate negative transfer.

**5.6 Acceleration**

Training spatio-temporal graph neural networks is a challenge especially on graph with billions of parameters like social networks. Acceleration research can help achieve faster training for the models in addition to reducing the computational complexity of models. In [38],Zhou et al. attempted to accelerate the spatio-temporal graph neural networks through TGL, a unified framework for large-scale offline training. The implementation included a design a Temporal-CSR data structure and a parallel sampler to efficiently sample temporal neighbors to form raining mini-batches. TGL achieved similar or better accuracy compared to other spatio-temporal GNNs with an average of 13× speedup. However, further advances in the field to speed up of training and decrease the computational complexity is still and urgent research stream especially with graph sizes increasing with the data hype.

# 6 CONCLUSION

Graph Neural Networks have gained huge interest in the past few years. These powerful algorithms expanded deep learning models to non-Euclidean space. However, GNNs are limited to static graph structures assumption which limits the Graph Neural Networks' performance when the data varies with time.Spatio-temporal Graph Neural Networks are extension of Graph Neural Networks that takes the time factor into account. In this survey, we conduct a comprehensive overview of spatio-temporal graph neural networks. First, we provide a taxonomy which groups spatio-temporal graph neural networks into two categories based on the method time variant is introduced. We also discuss a wide range of applications of spatio-temporal graph neural networks. Finally, we suggest future directions based on current open challenges in spatio-temporal GNNs.

# 7 REFERENCES

[1] Bruna, J., Zaremba, W., Szlam, A., & LeCun, Y. (2013). Spectral networks and locally connected networks on graphs. *arXiv preprint arXiv:1312.6203*.

[2] Kipf, T., & Welling, M. (2017). Semi-Supervised Classification with Graph Convolutional Networks. ArXiv, abs/1609.02907.

[3] Veličković, P., Cucurull, G., Casanova, A., Romero, A., Lio, P., & Bengio, Y. (2017). Graph attention networks. *arXiv preprint arXiv:1710.10903*.

[4] Xu, K., Hu, W., Leskovec, J., & Jegelka, S. (2018). How powerful are graph neural networks?. *arXiv preprint arXiv:1810.00826*.

[5] Yun, S., Jeong, M., Kim, R., Kang, J., & Kim, H. J. (2019). Graph transformer networks. *Advances in neural information processing systems*, *32*.

[6] Yu, B., Yin, H., & Zhu, Z. (2017). Spatio-temporal graph convolutional networks: A deep learning framework for traffic forecasting. arXiv preprint arXiv:1709.04875.

[7] Cao, D., Wang, Y., Duan, J., Zhang, C., Zhu, X., Huang, C., ... & Zhang, Q. (2020). Spectral temporal graph neural network for multivariate time-series forecasting. Advances in neural information processing systems, 33, 17766-17778

[8] Simeunović, J., Schubnel, B., Alet, P. J., & Carrillo, R. E. (2021). Spatio-temporal graph neural networks for multi-site PV power forecasting. IEEE Transactions on Sustainable Energy, 13(2), 1210-1220.

[9] Chen, C., Li, K., Teo, S.G., Zou, X., Wang, K., Wang, J., & Zeng, Z. (2019). Gated Residual Recurrent Graph Neural Networks for Traffic Prediction. AAAI.

[10] Wu, Z., Pan, S., Long, G., Jiang, J., Chang, X., & Zhang, C. (2020, August). Connecting the dots: Multivariate time series forecasting with graph neural networks. In Proceedings of the 26th ACM SIGKDD international conference on knowledge discovery & data mining (pp. 753-763).

[11] Karimi, A. M., Wu, Y., Koyuturk, M., & French, R. H. (2021). Spatiotemporal Graph Neural Network for Performance Prediction of Photovoltaic Power Systems. Proceedings of the AAAI Conference on Artificial Intelligence, 35(17), 15323-15330. Retrieved from https://ojs.aaai.org/index.php/AAAI/article/view/17799

[12] Huang, L., Wu, L., Zhang, J., Bian, J., & Liu, T. Y. (2022). Dynamic Relation Discovery and Utilization in Multi-Entity Time Series Forecasting. arXiv preprint arXiv:2202.10586.

[13] Kan, J., Hu, K., Hagenbuchner, M., Tsoi, A. C., Bennamoun, M., & Wang, Z. (2022). Sign Language Translation with Hierarchical Spatio-Temporal Graph Neural Network. In Proceedings of the IEEE/CVF Winter Conference on Applications of Computer Vision (pp. 3367-3376).

[14] Oreshkin, B. N., Amini, A., Coyle, L., & Coates, M. (2021, May). FC-GAGA: Fully connected gated graph architecture for spatio-temporal traffic forecasting. In Proceedings of the AAAI Conference on Artificial Intelligence (Vol. 35, No. 10, pp. 9233-9241).

[15] Kim, B. H., Ye, J. C., & Kim, J. J. (2021). Learning dynamic graph representation of brain connectome with spatio-temporal attention. Advances in Neural Information Processing Systems, 34, 4314-4327.

[16] Chu, P., Wang, J., You, Q., Ling, H., & Liu, Z. (2021). TransMOT: Spatial-Temporal Graph Transformer for Multiple Object Tracking. ArXiv, abs/2104.00194.

[17] Li, Y., & Moura, J. M. (2019). Forecaster: A graph transformer for forecasting spatial and time-dependent data. arXiv preprint arXiv:1909.04019.

[18] Nicolicioiu, A.L., Duta, I., & Leordeanu, M. (2019). Recurrent Space-time Graph Neural Networks. NeurIPS.

[19] Kapoor, A., Ben, X., Liu, L., Perozzi, B., Barnes, M., Blais, M., & O'Banion, S. (2020). Examining covid-19 forecasting using spatio-temporal graph neural networks. arXiv preprint arXiv:2007.03113.

[20] Roy, A., Roy, K. K., Ali, A. A., Amin, M. A., & Rahman, A. M. (2021, July). Unified spatio-temporal modeling for traffic forecasting using graph neural network. In 2021 International Joint Conference on Neural Networks (IJCNN) (pp. 1-8). IEEE.

[21] Zhang, X., Huang, C., Xu, Y., Xia, L., Dai, P., Bo, L., ... & Zheng, Y. (2021, May). Traffic flow forecasting with spatial-temporal graph diffusion network. In Proceedings of the AAAI conference on artificial intelligence (Vol. 35, No. 17, pp. 15008-15015).

[22] Shen, Y., Li, L., Xie, Q., Li, X., Xu, G. (2022). A Two-Tower Spatial-Temporal Graph Neural Network for Traffic Speed Prediction. In: Gama, J., Li, T., Yu, Y., Chen, E., Zheng, Y., Teng, F. (eds) Advances in Knowledge Discovery and Data Mining. PAKDD 2022. Lecture Notes in Computer Science(), vol 13280. Springer, Cham. https://doi.org/10.1007/978-3-031-05933-9_32

[23] Wang, L., Adiga, A., Chen, J., Sadilek, A., Venkatramanan, S., & Marathe, M. (2022). CausalGNN: Causal-Based Graph Neural Networks for Spatio-Temporal Epidemic Forecasting. Proceedings of the AAAI Conference on Artificial Intelligence, 36(11), 12191-12199. https://doi.org/10.1609/aaai.v36i11.21479

[24] Li, Y. F., Gao, Y., Lin, Y., Wang, Z., & Khan, L. (2020). Time Series Forecasting Using a Unified Spatial-Temporal Graph Convolutional Network. In Proceedings of Preregister Workshop in 34th Conference on Neural Information Processing Systems.

[25] Shao, W., Jin, Z., Wang, S., Kang, Y., Xiao, X., Menouar, H., ... & Salim, F. (2022). Long-term Spatio-temporal Forecasting via Dynamic Multiple-Graph Attention. arXiv preprint arXiv:2204.11008.

[26] Hadou, S., Kanatsoulis, C. I., & Ribeiro, A. (2021). Space-time graph neural networks. arXiv preprint arXiv:2110.02880.

[27] https://aaafoundation.org/american-driving-survey-2014-2015/

[28] Yang, H. E.; French, R. H.; and Bruckman, L. S., eds. 2019.Durability and Reliability of Polymers and Other Materials in Photovoltaic Modules. Plastics Design Library. Amsterdam: Elsevier, William Andrew Applied Science Publishers, 1st edition edition. ISBN 978-0-12-811545-9. doi:10.1016/C2016-0-01032-X.

[29] Zheng, Hang & Ding, Xu & Wang, Yang & Zhao, Chong. (2021). Attention Based Spatial-Temporal Graph Convolutional Networks for RSU Communication Load Forecasting. 10.1007/978-3-030-92635-9_7.

[30] Bloemheuvel, S., Hoogen, J. V. D., Jozinović, D., Michelini, A., & Atzmueller, M. (2022). Multivariate Time Series Regression with Graph Neural Networks. arXiv preprint arXiv:2201.00818.

[31] Wen, Z., & Fang, Y. (2022, April). TREND: TempoRal Event and Node Dynamics for Graph Representation Learning. In Proceedings of the ACM Web Conference 2022 (pp. 1159-1169).

[32] Cummings, D., Brahmaroutu, A., Nassar, M., & Ahmed, N.K. (2022). Technology Growth Ranking Using Temporal Graph Representation Learning.

[33] Haifeng Chen, Yifan Deng, Shiwen Cheng, Yixuan Wang, Dongmei Jiang, and Hichem Sahli. 2019. Efficient Spatial Temporal Convolutional Features for Audiovisual Continuous Affect Recognition. In Proceedings of the 9th International on Audio/Visual Emotion Challenge and Workshop (AVEC '19). Association for Computing Machinery, New York, NY, USA, 19–26. https://doi.org/10.1145/3347320.3357690

[34] Bhattacharya, U., Mittal, T., Chandra, R., Randhavane, T., Bera, A., & Manocha, D. (2020, April). Step: Spatial temporal graph convolutional networks for emotion perception from gaits. In Proceedings of the AAAI Conference on Artificial Intelligence (Vol. 34, No. 02, pp. 1342-1350).

[35] Wang, Y., Cai, Y., Liang, Y., Ding, H., Wang, C., Bhatia, S., & Hooi, B. (2021). Adaptive Data Augmentation on Temporal Graphs. In Advances in Neural Information Processing Systems (pp. 1440–1452). Curran Associates, Inc.

[36] Lou, G., Liu, Y., Zhang, T., & Zheng, X. (2021). STFL: A temporal-spatial federated learning framework for graph neural networks. arXiv preprint arXiv:2111.06750.

[37] Panagopoulos, G., Nikolentzos, G., & Vazirgiannis, M. (2021, May). Transfer graph neural networks for pandemic forecasting. In Proceedings of the AAAI Conference on Artificial Intelligence (Vol. 35, No. 6, pp. 4838-4845).

[38] Zhou, H., Zheng, D., Nisa, I., Ioannidis, V., Song, X., & Karypis, G. (2022). TGL: A General Framework for Temporal GNN Training on Billion-Scale Graphs. arXiv preprint arXiv:2203.14883.